\def\year{2020}\relax
\documentclass[letterpaper]{article} % DO NOT CHANGE THIS
\usepackage{aaai20}  % DO NOT CHANGE THIS
\usepackage{times}  % DO NOT CHANGE THIS
\usepackage{helvet} % DO NOT CHANGE THIS
\usepackage{courier}  % DO NOT CHANGE THIS
\usepackage[hyphens]{url}  % DO NOT CHANGE THIS
\usepackage{graphicx} % DO NOT CHANGE THIS
\usepackage{graphicx}  % DO NOT CHANGE THIS
\usepackage{amssymb}
\usepackage{amsmath} 
\usepackage{mathtools}
\usepackage{todonotes}
\usepackage{comment}
\usepackage{algorithm}
\usepackage{subcaption}% http://ctan.org/pkg/subcaption
\DeclareCaptionSubType*{algorithm}
\usepackage[noend]{algpseudocode}
\MakeRobust{\Call}
\usepackage{xcolor}

\newcommand{\sS}{\mathcal{S}} % set of States
\newcommand{\sA}{\mathcal{A}} % set of Actions
\newcommand{\bE}{\mathbb{E}} % Expectation
\newcommand{\bR}{\mathbb{R}} % Reals
\newcommand{\citet}[1]{\citeauthor{#1}~\shortcite{#1}}
\newcommand{\citep}{\cite}

\title{Planning with Abstract Learned Models\\While Learning Transferable Subtasks}
\author{John Winder,\textsuperscript{\rm 1} Stephanie Milani,\textsuperscript{\rm 2} Matthew Landen,\textsuperscript{\rm 3} Erebus Oh,\textsuperscript{\rm 1}\\ \Large \textbf{Shane Parr,\textsuperscript{\rm 4} Shawn Squire,\textsuperscript{\rm 1} Marie desJardins,\textsuperscript{\rm 5} and Cynthia Matuszek\textsuperscript{\rm 1}}\\ % All authors must be in the same font size and format. Use \Large and \textbf to achieve this result when breaking a line
 \textsuperscript{\rm 1}University of Maryland, Baltimore County, 
 \textsuperscript{\rm 2}Carnegie Mellon University, 
 \textsuperscript{\rm 3}Georgia Institute of Technology, \\
 \textsuperscript{\rm 4}University of Massachusetts Amherst,
 \textsuperscript{\rm 5}Simmons University\\ %If you have multiple authors and multiple affiliations
 } 

\begin{document}
\maketitle
\begin{abstract}
We introduce an algorithm for model-based hierarchical reinforcement learning to acquire self-contained transition and reward models suitable for probabilistic planning at multiple levels of abstraction. We call this framework Planning with Abstract Learned Models (PALM). By representing subtasks symbolically using a new formal structure, the lifted abstract Markov decision process (L-AMDP), PALM learns models that are independent and modular. Through our experiments, we show how PALM integrates planning and execution, facilitating a rapid and efficient learning of abstract, hierarchical models. We also demonstrate the increased potential for learned models to be transferred to new and related tasks.
\end{abstract}

%%%%%%%%%%%%%%%%%%%%%%%%%%%%%%
%% Introduction             %%
%%%%%%%%%%%%%%%%%%%%%%%%%%%%%%
\section{Introduction}
%\todo[inline]{0.75--1pg ; total 7 pages +1 for refs}

\noindent Model-based reinforcement learning (RL) acquires a model of an agent's interaction with its surroundings. In this set of approaches, the learned model captures the stochastic effects of actions on future states and rewards. Once a model has been acquired, an agent can use it for probabilistic planning to anticipate behaviors that will lead to a well-chosen path through state space.

For challenging tasks in a complex environment, one common technique is to ``divide and conquer,'' breaking down an overall task into a hierarchy of separate, smaller \textit{subtasks} that are easier to solve and, ideally, reusable. Such subtasks are commonly represented and variously discussed in RL literature as \textit{skills}~\cite{konidaris2018skills} and \textit{options}~\cite{sutton1999between}.

However, in practice, hierarchical reinforcement learning (HRL) often relies on an expert's manual encoding of domain knowledge to design hierarchies of subtasks. These methods directly encode a designer's preconceptions and bias, due to the specification of subtasks' structure and scope.
For example, HRL techniques such as MAXQ~\cite{dietterich2000hierarchical}, HAM~\cite{parr1998reinforcement,bai2017efficient} and AMDP hierarchies~\cite{gopalan2017planning}, as well as most work on options, depend on experts specifying an interlocking hierarchy of subtasks. Methods that do learn structure, such as MLSH~\cite{frans2017meta} and HIRO~\cite{nachum2018data}, still predetermine properties like the depth and width of the hierarchy. 
While option/skill discovery research attempts to address these limitations, many assumptions currently remain, such as ad hoc heuristics for transfer mappings, how state abstraction functions are acquired~\cite{macglashan2015grounding}. As a result, it remains a significant issue that learned subtasks are specific to the context in which they are learned, both the training environment and the hierarchy in which they are embedded.

The ultimate aim of this research is to develop a method by which an agent can autonomously construct a robust hierarchy of subtasks and their models. To support planning and task transfer, these subtasks should: (1) be \textit{independent} of one another, meaning that each subtask is a complete Markov decision process unto itself, solvable without relying on other subtasks; and (2), be \textit{modular}, meaning that they can be swapped in or out of a hierarchy. Intuitively, these constraints encode the requirement that subtasks can be added and removed without breaking the learned hierarchy and can be transferred to new tasks.

We introduce Planning with Abstract Learned Models (PALM), a novel method for assembling task components into subtasks with self-contained, local models. When deployed to a new task, PALM computes each subtask's model while preserving its independence. The main contributions of this work are as follows:

\begin{enumerate}
    \item \textbf{Lifted Abstract Markov Decision Process} (L-AMDP): an independent, modular subtask representation.
    \item \textbf{PALM \textsc{Phase-2}}: a process for converting task hierarchies learned from demonstrations into lifted AMDPs.
    \item \textbf{PALM \textsc{Phase-3}}: an algorithm for acquiring transition models for all useful subtasks in a hierarchy via integrated planning and learning.
\end{enumerate}

%%%%%%%%%%%%%%%%%%%%%%%%%%%%%%
%% Background, Related Work %%
%%%%%%%%%%%%%%%%%%%%%%%%%%%%%%
\section{Background \& Related Work}

% Reinforcement learning
Our work on learning hierarchies of independent, modular subtasks is in the general research area of reinforcement learning (RL). An RL scenario considers an agent that executes actions determined by some policy, receives feedback (reward), observes the change in its world state, and updates its policy to increase the expected reward.

Such scenarios are defined by Markov decision processes (MDPs), $M = \langle \sS, \sA, T, R, \gamma \rangle$, comprised of: a set of states $\sS$; a set of actions $\sA$; a transition probability distribution $T : \sS \times \sA \times \sS \rightarrow [0, 1]$; a reward function $R : \sS \times \sA \times \sS \rightarrow \bR$; and a scalar discount factor $\gamma \in (0,1]$, which governs the importance of future rewards. The agent's goal is to find the most rewarding behavior over time, as represented by a probabilistic policy $\pi : \sS \times \sA \rightarrow [0, 1]$. That policy is often learned through the computation of a value function $V^\pi$, which captures the long-term value of being in a given state. Value is computed recursively over rewards received from following the policy through state-action space as $V^\pi(s) = \bE_\pi[\sum_{k=0}^{\infty} \gamma^k  r_{t+k+1} | s = s_t], \forall s \in \sS$.
The optimal policy $\pi^*$ maximizes cumulative expected future rewards, corresponding directly to an optimal value function.

% Hierarchical Reinforcement Learning
Hierarchical RL (HRL) is commonly used in situations that require agents to perform repeated sequences of actions over an extended period.
HRL decomposes a hard task into subtasks that are more focused, more manageable, and (ideally) repeatable.
% As an example, consider the task of collecting a key and unlocking a door with it: this could be break down into three subtasks, one for moving to an object, one for picking up an object, and one for interacting with an object.
Two notable HRL techniques are options~\cite{sutton1999between} and MAXQ~\cite{dietterich2000hierarchical}. Both accomplish \textit{temporal} abstraction, in which the agent reasons over actions that unfold over many time-steps rather than discrete, atomic actions.
HRL is also conducive to \textit{state} abstraction, in which similar grounded states are aggregated~\cite{boutilier1994using,abel2018state}. 
Combined, these make HRL well suited to address many open challenges in RL, namely mitigating the curse of dimensionality by reducing the state-action space, facilitating generalization across similar (abstract) states, and leveraging transfer learning across tasks.

\begin{figure}
    \centering
    \begin{subfigure}[t]{0.40\textwidth}
        \includegraphics[width=1.0\textwidth]{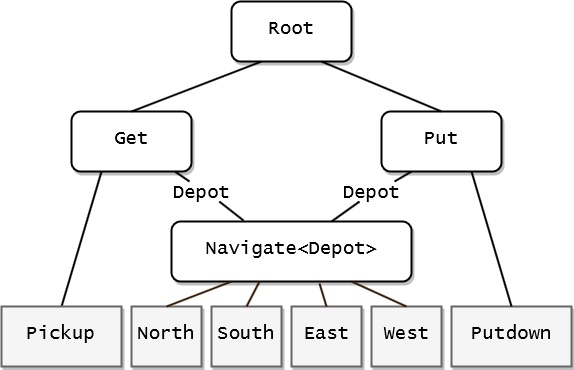}
        \caption{PALM-ET.}
    \label{fig:amdp-taxi-expert} 
    \end{subfigure}
    \par\bigskip
    \begin{subfigure}[t]{0.40\textwidth}
        \centering
        %\fbox{
        \includegraphics[width=0.80\textwidth]{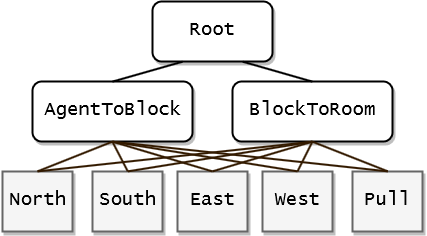}
        %}
        \caption{PALM-EC.}
        \label{fig:amdp-cleanup-expert}
    \end{subfigure}
    \caption{PALM-ET uses the classic expert Taxi hierarchy from \citet{dietterich2000hierarchical}, parameterized for multiple passengers. PALM-EC is our expert hierarchy for Cleanup.}
\end{figure}

% Hierarchy Learning algorithms
Central to HRL is the notion of a task \textit{hierarchy}, a graphical structure that encodes how subtasks and primitive actions relate (for example, what subtasks use other subtasks or actions).
Thus, each subtask may be viewed as a node in a directed acyclic graph (the hierarchy).
MAXQ, for example, decomposes an MDP into a set of smaller MDPs -- one per subtask -- with transitions and rewards derived from the base MDP.
It then computes the overall value function recursively down through each possible branch, using a piece-wise completion function to determine the expected discounted reward contributed by each subtask.

A drawback in much HRL research is the reliance on humans to design the task hierarchy. 
As a result, there has been significant research on autonomously learning such hierarchies.
%Explicit hierarchy learning algorithms begin with HEXQ~\cite{hengst2002discovering,hengst2004model}, in which an MDP is segmented into sub-MDPs (subtasks) by identifying boundary ``exit'' states.
Prior efforts in learning hierarchical structures of tasks include discovering sub-MDPs divided by suitable boundary ``exit'' states~\cite{hengst2002discovering,hengst2004model}; acquiring transferable option policies directly \cite{brunskill2014pac,topin2015portable};
learning high-level skills and abstract actions from demonstrations~\cite{konidaris2018skills}; learning hierarchies based on associating actions with relevant state features via CSRL~\cite{li2017efficient}; and, causally-annotated approaches that produce MAXQ-style task graphs, such as VISA~\cite{jonsson2005causal,vigorito2010intrinsically}, HI-MAT~\cite{mehta2008automatic}, and its extension,
HierGen~\cite{mehta2011hierarchical}.
% Discuss HAMQ-INT ?
% Discuss ARM-HSTRL?

While we restrict our examination in this paper to traditional, symbolic RL, there is a rich and growing body of work in the context of deep HRL~\cite{kulkarni2016hierarchical}.
MLSH~\cite{frans2017meta} learns master- and sub-policies jointly, with the former acting as a controller over the latter group.
The HIRO agent~\cite{nachum2018data} creates two tiers of policies, in which the lower level interacts with the grounded MDP based on goals directed by the higher level.
% This structure allows HIRO to greatly improve sample efficiency over similar methods, such as feudal networks, variational information maximizing exploration, and option-critic agents.
Notably, HIRO eschews any goal representation, so it lacks an explicit notion of subtask.
Along this thread, a HAC agent~\cite{levy2019learning} \textit{does} specifically consider goals and subtasks.
HAC combines hindsight experience replay with universal function approximators in an algorithm that learns, in parallel, multiple levels of policies.
% HAC outperforms HIRO on several robotic control domains, better handling cases of sparse reward.
Finally, the unified model-free HRL algorithm~\cite{rafati2019learning} combines intrinsic motivation with aspects of both skill and subgoal discovery.

One drawback of existing hierarchical methods, such as planning with (predefined) AMDP hierarchies~\cite{gopalan2017planning}, is the requirement of human intervention or curation at some (or many) parts of the design.
Examples include defining how subtasks chain together, picking the number or depth of subtasks, providing the low-level controllers beforehand.
This work is motivated by the goal of learning subtasks, both their relations among each other and their internal model dynamics, without expert knowledge.
While several HRL techniques can suffer from nonstationarity issues arising due to learning multiple levels of subtasks~\cite{nachum2018data}, our technique is devised to counter the problem without an impact to performance.
Lastly, in our approach, PALM learns AMDP subtasks that are independent and modular.
As such, these AMDPs can be removed or added without impairing the functioning of other subtasks or the hierarchy as a whole.
Taken together, the method we introduce facilitates the inclusion of new skills on transfer to different and more difficult tasks.

We draw a connection between PALM and an existing approach in planning and RL focused on abstract modular and independent tasks.
PLANQ-learning~\cite{grounds2005combining} synthesizes a hierarchical variant of Q-learning with STRIPS planning~\cite{fikes1971strips} to reason further into the future over a sequence of subtasks.
PPSB~\cite{segovia288planning}, planning with partially specified behaviors, builds on PLANQ with formalisms to afford more modular subtask decomposition.
PLANQ and PPSB encode preconditions and rules that comprise a knowledge base to inform per-subtask policies that are then learned independently, as guided by high-level plans.
PPSB is similar to our approach, though instead of AMDPs, PPSB relies on the partial specifications of abstract actions based in STRIPS. 
This framing requires expert-created conditions and operators for actions, including the models of the abstract actions.
PALM and L-AMDPs do not encode such information beforehand, except incidentally through state abstraction, which we show  can itself be learned from data without human supervision, due to the generality of AMDPs.
PPSB also assumes no  dead-ends and determinism within subtasks.
In contrast, PALM's recursive algorithm is designed to be tolerant to stochasticity and accounts for subtask failures explicitly.
On encountering a failed subtask, PALM escapes out and recurses up to replan at the appropriate parent AMDP.

%%%%%%%%%%%%%%%%%%%%%%%%%%%%%%
%% Approach                 %%
%%%%%%%%%%%%%%%%%%%%%%%%%%%%%%
\section{Approach}

PALM aims to remedy issues of existing techniques in a three-phase framework.
In \textsc{Phase-1}, our approach first relies on input sample demonstrations  and a hierarchy learner, producing a MAXQ-like hierarchy.
Our contributions begin in \textsc{Phase-2} with a procedure for creating a new subtask representation, the L-AMDP, from those extracted in \textsc{Phase-1}.
\textsc{Phase-3} then describes our novel algorithm for learning those subtask models in a hierarchy simultaneously at multiple levels, while preserving their self-contained properties.

\subsection{Learning from Demonstrations}
\textsc{Phase-1} uses existing HRL techniques to learn a MAXQ-style task hierarchy from demonstrations. 
% \todo{Maybe insert a sentence or two about MAXQ, either here or when it's mentioned in related work; it's pretty critical to understanding the paper and AAAI is a broad audience.} % John: I added it earlier
MAXQ decomposes a task hierarchically across subtasks, computing the value function by defining completion function to represent the amount of expected discounted reward recursively credited down through each branch of a task's subtasks.
Hierarchy-learning algorithms are typically assumed to have access to sample trajectories of an agent exploring and solving the domain (e.g., state-action-reward transition tuples), labeled with, minimally, whether the sample is a success or failure.
% The solution trajectories can be gathered from random walks in MDP instances pulled from a distribution for a given task, or from the paths taken by RL algorithms during training.

Given this set of solution demonstrations, $\mathcal{D}$, \textsc{Phase-1} takes any general existing algorithm $\mathtt{H}$ that assembles a MAXQ-style task graph, $\mathtt{H}(\mathcal{D}) \rightarrow G$.
This phase assumes the use of factored state space MDPs, such as states that are defined simply in terms of a feature vector.
There exist many possibilities for $\mathtt{H}$~\cite{mehta2011hierarchical,li2017efficient,levy2019learning}; 
%include HierGen~\cite{mehta2011hierarchical}, %CSRL~\cite{li2017efficient}, and HAC~\cite{levy2019learning}; 
we use HierGen~\cite{mehta2011hierarchical} because it creates a per-subtask state abstraction mapping.
Alternatively, \textsc{Phase-1} can be skipped if an expert supplies the task graph $G$ (a hierarchy of MAXQ subtasks).

\subsection{Assembling Lifted Abstract MDP Subtasks}
Given an expert or learned task graph from the previous phase, we present a novel process to convert all of its subtasks into a more general abstract MDP representation, \textbf{\textsc{Phase-2}} of the process. We refer to this novel intermediate representation of an abstract subtask as \textbf{the Lifted AMDP}, or L-AMDP.
At a high level, \textsc{Phase-2} maps the task graph $G$ to $H$, a hierarchy of Lifted AMDPs.

We depart from standard approaches by representing subtasks abstractly as decision problems complete unto themselves.
To do so, we redefine subtasks using the formalization of abstract MDPs (AMDPs)~\cite{gopalan2017planning}.
AMDPs differ from the more familiar framing of subtasks as skills or options in that each AMDP subtask has its own \textit{local} model of reward and transitions, from which value functions and policies may be generated. 

Formally, an AMDP subtask is an MDP with abstract components, $\tilde{M}$ = $\langle \tilde{\mathcal{S}}, \tilde{\mathcal{A}}, \tilde{T}, \tilde{R}, \gamma, \tilde{\mathcal{G}}, \tilde{\mathcal{F}}, \phi \rangle$, consisting of: abstract actions $\tilde{\mathcal{A}}$, comprised of child subtasks (e.g., other AMDPs or primitive actions from $M$); sets of terminal goal and failure states $\tilde{\mathcal{G}},\tilde{\mathcal{F}} \subset \tilde{\mathcal{S}}$; and (optionally) state abstraction $\phi : \mathcal{S}\rightarrow\tilde{\mathcal{S}}$,
which maps ground states to abstract state space by aggregating similar states or removing features.

Extending this concept, \textit{lifted} AMDPs are parameterized over features (state factors).
An L-AMDP contains a goal predicate $\tau$, a failure predicate $\chi$, subtask parameters $\theta$ (if any), and a state abstraction function $\phi$.
We formally define an L-AMDP as $\tilde{M}_\theta$ = $\langle\cdot,\tilde{\mathcal{A}},\cdot,\tilde{R}\sim\{\tau,\chi\},\cdot, \tilde{\mathcal{G}}\sim\{\tau\}, \tilde{\mathcal{F}}\sim\{\chi\},\phi \rangle$.
On grounding to a target task MDP, they serve as AMDP subtasks with models missing; the remaining components are induced or learned during \textsc{Phase-3}.

\textsc{Phase-2} is a procedure that builds a hierarchy of Lifted AMDPs, $H$, from a task graph $G$.
Construction of L-AMDPs occurs one at a time, following a reverse topological ordering of $G$.
Intuitively, \textsc{Phase-2} first wraps primitive actions (leaf nodes of $G$) into ``subtasks,'' then creates one L-AMDP subtask for each composite action (internal node of $G$), and ends by making a subtask for the root. There is always a root subtask, with terminal conditions identical to the task MDP.

% For each L-AMDP assembled, \textsc{Phase-2} creates or learns the following components.
Given our assumption of factored states, each subtask either already possesses terminal conditions. Or, we can train a classifier to learn goal and failure predicates based on terminal states in $\mathcal{D}$. $\tau$ and $\chi$ differ from the termination predicate in MAXQ by capturing success and failure separately.
This property of L-AMDPs is crucial for $\tilde{R}$, the induced pseudo-reward function. 
We make the standard assumption that subtasks adhere to a goal-fail structure. 
In this scheme, the highest reward is received at a goal state, the most negative reward is received at any failure state, and a default reward is otherwise observed; this assumption simplifies and bounds the shape of $\tilde{R}$.
Although $\tilde{R}$ has a known shape, a condition-based pseudo-reward may be included as well, either learned from $\mathcal{D}$, or added manually.
Thus $\tilde{R}$, like $\tilde{T}$, once grounded, must be learned to a task MDP in the next phase, during execution.
%$\tilde{R}$ is determined by $\tau$ and $\chi$, giving +1.0 or -1.0 for success and failure, respectively.
%A constant step-wise pseudo-reward may also be applied (normalized between [-1.0, 1.0]); condition-based rewards must be added manually.

We use the (optional) $\theta$ list of parameterizable features.
A parameterized L-AMDP may generate multiple realized AMDPs, one for each possible grounding in the target task. For example, the Taxi hierarchy's \textsc{Navigate} subtask (Figure~\ref{fig:amdp-taxi-expert}) may be grounded once per destination depot in a given MDP.
HierGen~\cite{mehta2011hierarchical} provides a straightforward way of generating $\theta$ based on what factors are checked or change in the trajectories ($\mathcal{D}$) for that subtask (e.g., the positional coordinates of depots in Taxi).
Similarly, the state abstraction may be hand-engineered or learned, such that $\phi$ includes all features from $\tau$, $\chi$, and $\theta$.
Finally, \textsc{Phase-2} makes $\tilde{\mathcal{A}}$ by linking to all AMDPs that correspond to child subtasks of this subtask as defined in $G$; reverse topological ordering ensures these AMDPs already exist.
The missing $\gamma$ will likewise be inherited from the ground MDP.

In summary, the Lifted AMDP is a novel, modular, independent subtask representation. In \textsc{Phase-2}, each input MAXQ task node is converted into an L-AMDP.

\paragraph{Subtask Independence.}
The L-AMDP subtask representation devised for PALM leads to independently computable models. In possessing each component of an MDP, an AMDP is itself an MDP, and can be solved using any standard approach.
Deployed to some task (\textsc{Phase-3}), whenever the agent first executes an L-AMDP subtask, it is automatically grounded to its particular circumstance, with the missing pieces induced or learned gradually through experience.
Computing plans inside a grounded AMDP subtask is therefore self-contained: it may be solved as with any MDP.

This is an improvement over dependencies present in both options and MAXQ subtasks.
Computing the hierarchical value function for a MAXQ subtask requires decomposing via the completion function into its child subtasks.
While this in principle reduces the size of the value function, it acquires this property by entangling subtasks with ones below it.
R-MAXQ~\cite{jong2008hierarchical}, the model-based extension of MAXQ, highlights the complexity of learning MAXQ subtask models: its recursive variant of the Bellman equation must consider the stochastic transitions across \textit{all} child subtasks, for all their possible terminal states.
For options, approximating an option's multi-time model requires knowing all of the (possibly infinite) number of time-steps across its subtasks \cite{abel2019expected}. 
In general, computing option plans in complex hierarchies requires solving how their models compose jointly, in a dependence akin to that of the MAXQ completion function~\cite{silver2012compositional}.
The practical effect of L-AMDP subtask independence is that, on grounding, subtasks may be computed directly, in isolation, without any recursive decomposition.

\paragraph{Subtask Modularity.}
L-AMDPs also improve the modularity of subtasks: unnecessary subtasks can easily be pruned, and new ones incorporated.
For example, imagine transferring from the classic Taxi task to a new task that requires refueling.
The existing hierarchy (Figure~\ref{fig:amdp-taxi-expert}), defined in terms of L-AMDPs, can be transferred directly; the learned models of its abstract subtasks are still valid, due to the AMDP's state abstraction and independent model.
All that needs to be added is a newly initialized \textsc{Refuel} subtask, included as a child of the \textsc{Root} subtask, with its set of abstract actions $\tilde{\mathcal{A}}$ containing \textsc{Navigate} and a primitive action for refueling (assuming it is available in the task MDP).

When a new subtask is added to a hierarchy, its presence would be included in any parent subtasks' abstract action set; only the new subtask and any parents need to be updated.
Assuming an optimism-under-uncertainty model-based scheme is used, the newly available transitions due to the new subtask are incentivized to be explored.
Removing a subtask is likewise straightforward: the subtask only needs to be removed from its parents' $\tilde{\mathcal{A}}$.
Any learned transitions involving that action would, thus, be disallowed or deleted.

\begin{algorithm}[t]
\begin{algorithmic}[1] 
	\Function{Start}{Hierarchy $H$, MDP $M$, State $s_0$}
    	\State Initialize models for all AMDPs $\tilde{M}_i \in H$
    	\State \Call{PALM}{$H$, $M$, \Call{Root-Index}{$H$}, $s_0$} 
    \EndFunction

    \Function{PALM}{$H$, $M$, AMDP index $i$, $s_t$}
    	\State $\langle \tilde{\mathcal{S}}, \tilde{\mathcal{A}}, \tilde{T}, \tilde{R}, \gamma, \tilde{\mathcal{G}}, \tilde{\mathcal{F}}, \phi \rangle \leftarrow \tilde{M}_i \leftarrow \Call{Node}{H,i}$ %\Comment{{Retrieve the AMDP}}
        \State $\tilde{s}_t \gets \phi(s_t)$ %\Comment{{Apply state abstraction to $s_t$}}
        \While{$\tilde{s}_t \notin \tilde{\mathcal{G}} \cup \tilde{\mathcal{F}}$} \Comment{{Execute until termination}}
            \State $\tilde{\mathcal{S}} = \tilde{\mathcal{S}} \cup \tilde{s_t}$
            \State $\pi \gets$ \Call{Plan}{$\tilde{M}_i, \tilde{s}_t$} \label{line:plan} \Comment{{Compute local policy}}
                 \State $a \gets \pi(\tilde{s}_t)$ \Comment{{Get next planned action, $a \in \tilde{\mathcal{A}}$}}
                \If{\Call{Is-Primitive}{$a$}} % NOTE: this does not 100% match the actual code in the AMDP / PALM repo, because it is trickier to include that final recursive base case in this recursive function pseudocode. So, basically the base case is elevated out of its final recursive call and defined here, in this location instead -- it is functionally identical.
                    \State $s_{t+1} \gets $\Call{Execute}{$M, a$} %\Comment{Execute, observe the next ground state (base case)}
                \Else
                    \State $j \gets$ \Call{Child}{$H, i, a$} %\Comment{{$a$ links $\tilde{M}_i$ to subtask $\tilde{M}_j$}}
                    \State $s_{t+1} \leftarrow $ \Call{PALM}{$H, M, j, s_t$} %\Comment{{Recurse to plan and execute the chosen subtask}}
                \EndIf
                \State $\tilde{s}_{t+1} \gets \phi(s_{t+1})$  %\Comment{{Abstract the next ground state}}
                \State $r \gets \tilde{R}(\tilde{s}_t,a,\tilde{s}_{t+1})$ \label{line:reward} %\Comment{{Obtain task's pseudo-reward}}
                \State \Call{Update-Model}{$\tilde{T}, \tilde{R}, \tilde{\mathcal{G}},  \tilde{\mathcal{F}}, \tilde{s}_t, a, \tilde{s}_{t+1}, r$} \label{line:update-model}
				% \If{\Call{Task-Completed}{$\tilde{M}_i,\tilde{s}_{t+1}$}} \label{line:task-completed}
                %      \State \Call{Handle-Success}{$\tilde{M}_i, \tilde{s}_t, a, \tilde{s}_{t+1}, r$}
                % \EndIf
                % \If{\Call{Task-Failed}{$\tilde{M}_i,\tilde{s}_{t+1}$}}
                 %     \State \Call{Handle-Failure}{$\tilde{M}_i, \tilde{s}_t, a, \tilde{s}_{t+1}, r$} \label{line:handle-failure}
                % \EndIf
                \State $t \leftarrow t+1$ %$\tilde{s}_t \leftarrow \tilde{s}_{t+1}$
			\EndWhile
        \State \Return $s_t$ \Comment{The current ground state}
    \EndFunction 
\end{algorithmic}
\caption{Planning with Abstract Learned Models} 
\label{alg:palm}
\end{algorithm}

\subsection{Integrating Planning and Learning}
\textbf{\textsc{Phase-3}} sees a new model-based hierarchical RL algorithm (Algorithm~\ref{alg:palm}) applying the hierarchy from \textsc{Phase-2} to a new (previously unseen) task MDP $M$.
This algorithm recursively integrates planning and learning to acquire its subtasks' models while solving $M$. We refer to the algorithm as PALM: Planning with Abstract Learned Models.

In model-based RL, the agent uses its observations to bootstrap an approximation of the model, $\tilde{T}$ and $\tilde{R}$, which express the probabilistic effect of actions on the environment.
In our case, this occurs at multiple levels of abstraction, with PALM learning the $\tilde{T}$ and $\tilde{R}$ of each subtask, updating the model after using it to make a decision. Conceptually, PALM plans the solution to a given task (a policy $\pi$), selects a subtask, recurses, and continues this process successively until reaching and executing a primitive action of $M$. Thus, planning and execution are interleaved.

The inputs of PALM are $M$, an initial ground state $s_0 \in \mathcal{S}$, and a hierarchy $H$ of L-AMDPs.
Planning begins at the root subtask.
With each recursive call to PALM, $s_t$ is projected into the given AMDP's state space by applying state abstraction.
PALM then computes a local policy $\pi$ for the AMDP, selecting the next planned action.
If that action is primitive ($a\in\mathcal{A}$), executing it in the base environment causes a ``real'' transition in the world, and it returns the next ground state, $s_{t+1}$.
Otherwise, PALM retrieves the linked child subtask in the hierarchy, and recurses down to it, repeating this process.
On completion of the execution step or recursive call, PALM abstracts the new ground state, $\tilde{s}_{t+1}$, obtains the pseudo-reward $r$ for the abstract transition, and updates the current AMDP's model, recomputing the approximation of $\tilde{T}$ and $\tilde{R}$ based on the observed transition. 
%For line~\ref{line:reward}, 
%The model-learning algorithm in each AMDP explicitly handles success and failure separately.
%For our implementation, in the case that $\tilde{s}_t$ is a goal or failure, $r$ is overridden to apply the maximum or minimum value, respectively.

It is desirable for PALM to be a general framework.
\textsc{Phase-3} is designed to permit the use of any MDP planner, such as Value Iteration, UCT~\cite{kocsis2006bandit} methods like PROST~\cite{keller2012prost}, or bounded RTDP~\cite{mcmahan2005bounded}.
PALM also supports any standard model-based approach such as $E^3$~\cite{kearns2002near} or R-MAX~\cite{brafman2002rmax}.
Additionally, because subtasks are independent, they may use different planners as appropriate---for example, if value function approximation is required in one subtask but not another.

\paragraph{Avoiding Nonstationarity.}
One common issue in HRL is that of nonstationarity: so long as a lower-level model has not converged to a stable point, it defines a moving target for any higher-level subtask relying on that model.
Learning on a nonstationary process upends the assumptions needed in RL to update via bootstrapping and guarantee optimality.
To avoid the effects of nonstationarity, we rely on the ``knows-what-it-knows'' (KWIK) framework.
A KWIK algorithm such as R-MAX~\cite{li2011knows,szita2011agnostic} reasons explicitly about known or unknown transitions.
We employ a strategy of solidifying lower subtask models first, then learning successively higher levels.
During execution, PALM simply returns a signal indicating the known/unknown status of the transition that just occurred, informing the parent task if it should ignore or process its own transition. 
In effect, subtask models are cemented before their parent models are learned.
Without this update strategy, we find empirically that PALM produces solutions swiftly while learning on nonstationary models, often achieving more optimal behavior; however, this strategy ensures optimality across levels of abstraction such that PALM converges to a recursively optimal policy.

\paragraph{Complexity.}
We consider the complexity of each AMDP subtask individually.
In the execution of PALM on a given AMDP (line~\ref{line:plan}), computational complexity is dominated by the planning algorithm.
In the worst case, this planner is recomputed at each step.
The sample complexity for an AMDP subtask in general will be $O(\rho)$, where $\rho$ is the sample complexity of the model-learning algorithm used; for R-MAX, which we use, $\rho = |\mathcal{S}|^2|\mathcal{A}|V_{max}^3\epsilon^{-3}(1-\gamma)^{-3}$ given $M$'s maximum value $V_{max}$ and the PAC-MDP parameter $\epsilon$ \cite{li2012sample}. 
R-MAX is a well-studied, PAC-MDP algorithm,
with guarantees of convergence and bounded space and sample complexity~\cite{strehl2009reinforcement}. 
Thus, hierarchies of fewer nodes and shallower depth are preferable in terms of sample complexity.
% Depending on both the planning algorithm and model-learning algorithm, it is possible to cache planned policies between iterations, greatly ameliorating the computational cost of planning at each abstract step.
% In our experiments, we omit caching when comparing PALM with R-MAXQ and enable it for the PALM-only experiments.

\section{Experimental Methodology}

We compare our method against R-MAXQ~\cite{jong2008hierarchical}, the most closely aligned existing algorithm to PALM and its aims.
R-MAXQ unites R-MAX with MAXQ, achieving a model-based approach of the latter method's value function decomposition technique by specifying a recursive variant of the Bellman operator.
As opposed to other HRL algorithms, R-MAXQ is similar to PALM in representing abstract models concretely; unlike PALM, R-MAXQ's models do not meet the criteria we desire for subtask independence and modularity.
To keep our PALM in line with R-MAXQ, we parallel it by using Value Iteration as the planner and R-MAX as the model-based RL algorithm for all subtasks; these are baselines, and due to subtask independence, more sophisticated techniques could be used as appropriate.

\begin{figure}
\centering
\begin{subfigure}[t]{0.18\textwidth}
    \centering
    \includegraphics[width=1.0\textwidth]{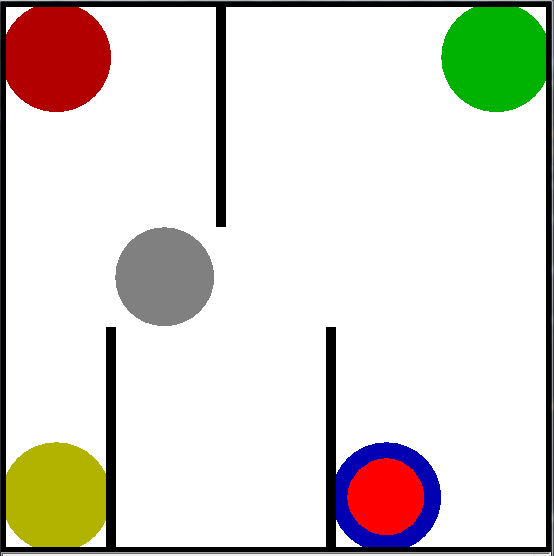}
\end{subfigure}
~~~~~ % Don't do this, it's hacky and gross
\begin{subfigure}[t]{0.18\textwidth}
    \centering
    \includegraphics[width=1\textwidth]{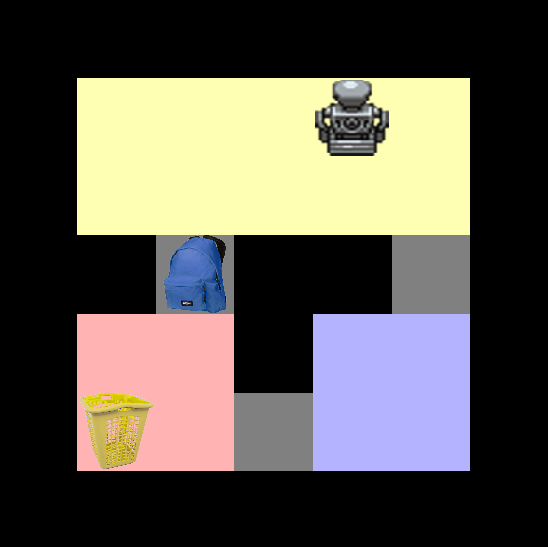}
\end{subfigure}
\caption{\label{fig:example-domains}Example starting states pulled from a distribution over possible tasks. We use the factored OO-MDP representation for tasks  \protect\cite{diuk2008object}. A Taxi task, left, contains the agent ``taxi'' (gray), walls, depots, and passenger (red). A Cleanup task, right, includes blocks, doors, rooms, and the agent ``robot'' facing north.}
\end{figure}

\paragraph{Domains.}
The \textbf{Taxi} domain~\cite{dietterich2000hierarchical} is a common HRL problem where the agent, a taxi, must collect passengers and ferry them to different destinations.
The \textbf{Cleanup} domain simulates a robot that tidies a house by putting blocks where they belong, similar to the game of Sokoban~\cite{macglashan2015grounding,guez2019investigation}.
The agent must navigate a grid-world composed of rooms and blocks, pushing and pulling blocks until one (or more) target blocks are in a room of a matching color.
Cleanup tasks appear deceptively simple.
They present a combinatorial explosion of state space as more blocks and rooms are added.
Moreover, agents frequently encounter edge cases and bottlenecks that are difficult to exit (e.g., blocks in a corner). 
Visualizations of a classic Taxi task and a Cleanup task are shown in Figure~\ref{fig:example-domains}. 

\begin{figure}[t]
    \centering
    \includegraphics[width=0.4\textwidth]{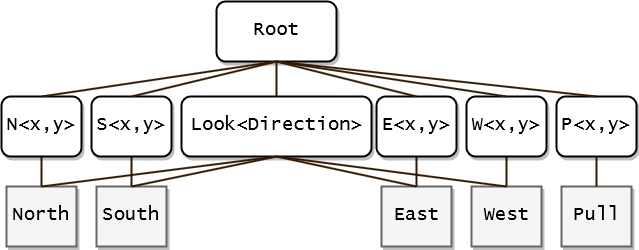}
    \caption{PALM-AC, amended by experts.}
    \label{fig:amdp-cleanup-advanced}
\end{figure}

\paragraph{Hierarchies.}
All PALM methods use a hierarchy of L-AMDP subtasks that, on each trial, are grounded to a new, random target MDP.
The lifted hierarchies we describe use an abbreviation scheme: \textbf{E} is an expert-designed hierarchy, \textbf{H} is a HierGen-learned hierarchy, \textbf{A} is an amended hierarchy that updates a learned hierarchy with expert knowledge, \textbf{T} is a hierarchy for Taxi, and \textbf{C} is a hierarchy for Cleanup.
Thus, PALM-\textbf{ET} and PALM-\textbf{HT} are the expert and learned hierarchies for the Taxi domain, following directly from \citet{dietterich2000hierarchical} and \citet{mehta2011hierarchical}, respectively.
PALM-\textbf{EC} (Figure~\ref{fig:amdp-cleanup-expert}), our expert hierarchy for Cleanup, decomposes the task in terms of moving to a block and then the block to a room.
PALM-\textbf{HC}, the learned Cleanup hierarchy, degenerates into a flat hierarchy without subtasks beyond \textsc{Root}, which contains only the primitive actions.
We find this negative result (from our use of HierGen) surprising, and discuss it further later.
Informed by HC, PALM-\textbf{AC} (Figure~\ref{fig:amdp-cleanup-advanced}), is an amended, second-draft expert hierarchy for Cleanup.
In AC, each primitive action is wrapped in an AMDP subtask parameterized by the x-y coordinates of the destination relative to the agent position.
Additionally, we define a \textsc{Look} task for turning to look in a direction (without hitting a wall), which promotes an agent's ability to plan to face a block before pulling it.
Combining these parameterized, shielded primitive actions with state abstraction means AC ignores all irrelevant features, and never plans an action that is illegal or results in a self-transition.

%%%%%%%%%%%%%%%%%%%%%%%%%%%%
%% Experimental Results   %%
%%%%%%%%%%%%%%%%%%%%%%%%%%%%
\section{Experimental Results}

We examine the usefulness of learned hierarchies in comparison with expert-specified hierarchies, and we consider the performance of PALM in terms of the cumulative steps taken and reward acquired across episodes. Our figures are shown with the 95\% confidence region shaded.

%
%   Taxi, R-MAXQ and PALM comparison
%
\begin{figure}[t]
	\centering
\begin{subfigure}[t]{0.23\textwidth}%{0.465\textwidth}
    \centering
    \includegraphics[width=1.0\textwidth]
    {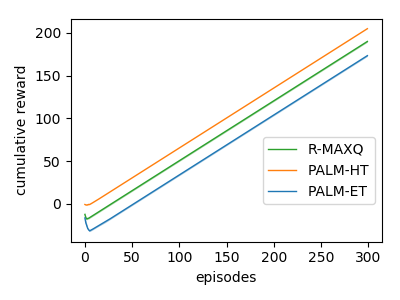}
    \caption{\label{fig:taxi-small-rmaxq-palmet-palmht} Small, deterministic Taxi.}
\end{subfigure}
\begin{subfigure}[t]{0.23\textwidth}
    \centering
    \includegraphics[width=1.0\textwidth]
    {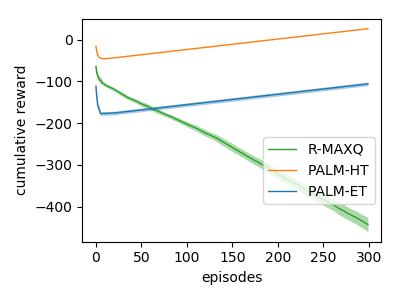}
    \caption{\label{fig:taxi-classic-rmaxq-palmet-palmht} Classic fickle Taxi.}
\end{subfigure}
  \caption{In \ref{fig:taxi-small-rmaxq-palmet-palmht}, without randomness, all approaches rapidly find an optimal policy (reflected in their asymptotic trends). In \ref{fig:taxi-classic-rmaxq-palmet-palmht}, ET and HT again converge quickly due to the ease of computing AMDP models relative to R-MAXQ's approach. HT consistently learns with fewer samples; while HT also learns fewer models, they are larger and more complex.}
  \label{fig:taxi-rmaxq-results}
\end{figure}

\subsection{Baseline Results}
Figure~\ref{fig:taxi-rmaxq-results} compares PALM with R-MAXQ.
In a small Taxi task ($1\times5$ grid) with deterministic transitions, R-MAXQ exceeds PALM when using the same hierarchical structure (ET).
In Figure~\ref{fig:taxi-classic-rmaxq-palmet-palmht},
we observe a different asymptotic relationship when facing stochasticity.
R-MAXQ struggles to compute the recursive models of its subtasks, even in this task with only 100 states.
Note that these results match those reported in \citet{jong2008hierarchical}, and belie the fact that R-MAXQ \textit{does} ultimately approximate correct models, just in episodes beyond those shown.

These results are even more stark in Cleanup, due to the exacerbation of R-MAXQ's scaling issues in handling larger, more complex state spaces.
R-MAXQ experimentally requires orders of magnitude more computation time than PALM for the same domain.
To highlight the stark difference in the actual time operations between the two methods, we report the empirical runtime difference of 20 trials on a small Cleanup task with $|\mathcal{S}|$ in the hundreds.
Using EC for both, PALM decreases from $100$ ms per episode to $10$ once its models have converged, while R-MAXQ consistently averages $10^4$ ms.\footnote{Performed on i7-4790K CPU @ 4.00 GHz, 20GB of RAM.}
Behaviorally, PALM quickly converges towards a near-optimal policy while R-MAXQ conducts excessive exploration.
We observe that R-MAXQ continually expands down branches of a hierarchy that are unhelpful to solving the overall goal, for example, continually re-entering a room and going to as many spaces as possible and facing in all directions.
This difference is due to subtask independence: unlike with MAXQ, planning with an AMDP hierarchy expands only those child subtasks actually used in the rollout of a hierarchical policy~\cite{gopalan2017planning}.
Thus, PALM's focused expansion learns models that solve the harder Cleanup tasks quickly, where R-MAXQ has superfluous branching.

%
%   Taxi Classic Fickle and Large results
%
\begin{figure}[t]
	\centering
\begin{subfigure}[t]{0.23\textwidth}
    \centering
    \includegraphics[width=1.0\textwidth]
    {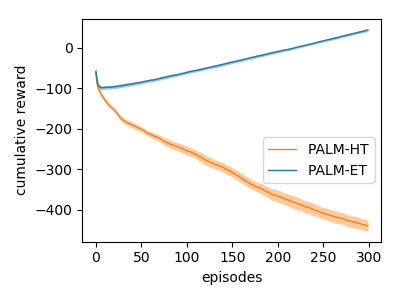}
    \caption{\label{fig:taxi-classic-fickle-2}Classic, 2 fickle passengers.}
\end{subfigure}
\begin{subfigure}[t]{0.23\textwidth}
    \centering
    \includegraphics[width=1.0\textwidth]
    {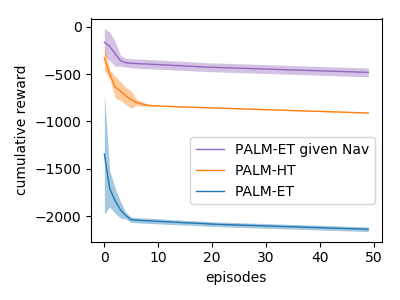}
    \caption{\label{fig:taxi-large}Large, 1 passenger.}
\end{subfigure}
  \caption{In \ref{fig:taxi-classic-fickle-2}, multiple fickle passengers cause HT to vacillate while learning mid-level models. In \ref{fig:taxi-large}, we provide a converged model of the navigation subtask to an ET agent, highlighting the benefit of transferring learned models.}
  \label{fig:taxi-fickle-results}
\end{figure}

%
%   Cleanup results
%
\begin{figure}[t]
    \centering
\begin{subfigure}[t]{0.23\textwidth}
  \centering
	\includegraphics[width=1.0\textwidth]{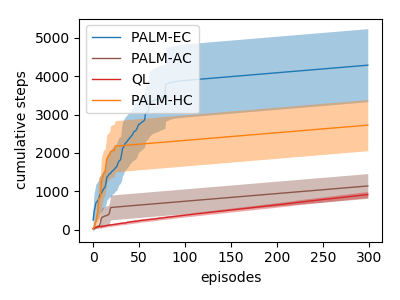}
	\caption{\label{fig:cleanup-1b-3r-7x5}1 block, 3 rooms, 5$\times$7 grid.}
\end{subfigure}
\begin{subfigure}[t]{0.23\textwidth}
    \centering
    \includegraphics[width=1.0\textwidth]
    {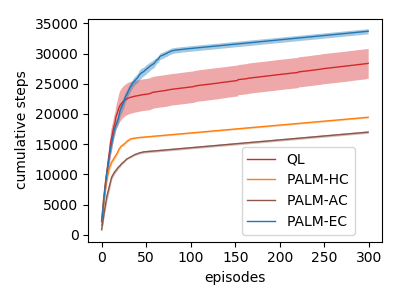}
    \caption{\label{fig:cleanup-1b-3r-7x7}1 block, 3 rooms, 7$\times$7 grid.}
\end{subfigure}
\par\bigskip
\begin{subfigure}[t]{0.23\textwidth}
    \centering
    \includegraphics[width=1.0\textwidth]
    {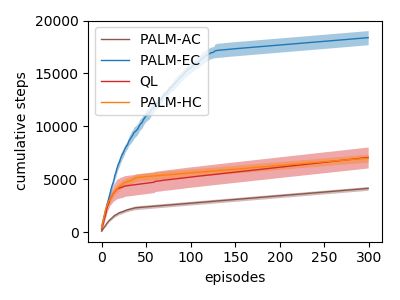}
    \caption{\label{fig:cleanup-2b-2r-5x5-1target}2 blocks (1 target), 5$\times$5 grid.}
\end{subfigure}
\begin{subfigure}[t]{0.23\textwidth}
    \centering
    \includegraphics[width=1.0\textwidth]
    {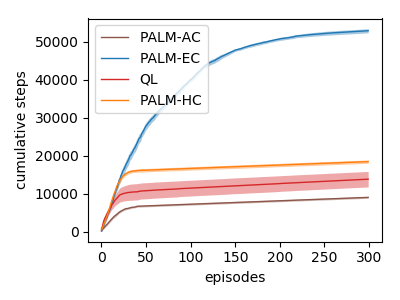}
    \caption{\label{fig:cleanup-2b-2r-5x5-2targets}2 blocks, 5$\times$5 grid.}
\end{subfigure}
\caption{Cumulative steps for the Cleanup domain, highlighting the number of samples needed to reach convergence.}
\label{fig:cleanup-results}
\end{figure}

\subsection{Results for Expert vs. Learned Hierarchies}

We now consider the impact of learning a hierarchy.  
Figure~\ref{fig:taxi-classic-rmaxq-palmet-palmht} shows how HT, a hierarchy learned from data without human supervision, can surpass the efficiency of one made by an expert, ET, with the former attaining roughly the same asymptotic performance as ET in only a fraction of the samples required.
PALM makes it possible to create everything needed for hierarchical planning via sampled experiences without expert knowledge or intervention.

In Figure~\ref{fig:taxi-fickle-results}, we examine hierarchies on increasingly more complex tasks.
However, ET achieves greater cumulative reward in the successive experiments than HT, with the latter faltering to increasingly greater extents as more passengers are added.
HT reasons about putting down passengers in its top-most AMDP, slowing it down when there are more than one, whereas ET more cleanly separates the process of retrieving and depositing passengers.
As we scaled to variants with three and four passengers, we found this trend continued to worsen for HT, while ET could achieve greater cumulative reward in the successive experiments, scaling more gracefully in contrast to the sinking asymptotic trend of HT.
For Figure~\ref{fig:taxi-large}, the large Taxi variant, on a 20$\times$20 grid with only one passenger, HT has the same lead over ET as in \ref{fig:taxi-classic-rmaxq-palmet-palmht}.

For Cleanup, in Figure~\ref{fig:cleanup-results}, we report results in cumulative steps rather than reward (each complete episode yields a reward of exactly 1.0) as complexity is varied among tasks.
Specifically, we consider when there are more rooms, Figures~\ref{fig:cleanup-1b-3r-7x5} and \ref{fig:cleanup-1b-3r-7x7}, or fewer rooms but more blocks, Figures~\ref{fig:cleanup-2b-2r-5x5-1target} and \ref{fig:cleanup-2b-2r-5x5-2targets}.
The key property to observe in Figure~\ref{fig:cleanup-results} is that the time-to-convergence signifies how effectively the algorithm computes its models, with the ``elbow'' of each trend indicating the point at which the models solidify.
Thus, we report cumulative instead of average steps; with the latter, it is harder to distinguish the differences in the long-term model-learning trends among the algorithms (typically, they all eventually have the same average number of steps once their models have converged).
We note that the asymptotic relations among the PALM methods hold across all tasks: EC requires the most samples, while HC is in the middle, and the redesigned AC solves the domains most readily.
Both HC and AC outperform QL, which we include to benchmark model-free learning (without planning).
In comparison, the original expert design, EC, is much less effective, taking statistically significantly more steps to learn than QL, despite having fewer models than AC.
AC is created based on our observations of EC and then HC; the fact this second iteration expert hierarchy performs better after observing a learned one highlights the value of human cooperation with learning algorithms in designing solutions.

We note a finding that emphasizes the challenge of generalization that endures for hierarchy-learning algorithms.
Specific to the algorithm we used, HierGen, we observed it would degenerate if trained on $\mathcal{D}$ of sufficient environmental complexity.
For example, if $\mathcal{D}$ contained two Cleanup MDPs with the same number of rooms and blocks but no shared features (e.g., colors or door locations), HierGen would produce only flat MAXQ graphs (causally-annotating subsequences into a precedence graph of solely one node).
Despite this degeneration, 
we remark that the resulting HC still produces valid, reasonably efficient solutions in $\textsc{Phase-3}$. 
In each case, it converges significantly faster than EC.

\subsection{Independence, Modularity, \& Transfer}

%%%% Main Idea: subtasks are independent, leading to a more flexible framework
The most significant difference between our formulation and R-MAXQ, and other subtask representations more generally, is that PALM yields independent models.
Practically, PALM circumvents the typical HRL approach of creating interdependent ``puzzle piece'' forms of temporal abstraction that must link together precisely.
For example, options must terminate where another can initiate, skills require chaining from target to target, and MAXQ does not distinguish between goal and failure termination predicates.
As a result, we identify a key benefit of PALM: graceful replanning on failure.
Should a subtask fail, control simply returns to its parent, and so on as needed, to the appropriate level for replanning, and a new plan is generated. 

We also highlight the transferability implications of subtask modularity: PALM can learn one subtask, then transfer its abstract models to new, related tasks, greatly accelerating overall performance.
To transfer in a model, it is first necessary to obtain either an expert-defined one or a learned one acquired via training on an MDP sampled from the same universe. 

We include an example of transfer in Figure~\ref{fig:taxi-large}, and refer to standard metrics for transfer in RL~\cite{taylor2009transfer}.
In this case, a converged model for the \textsc{Navigation} subtask is transferred to a hierarchy deployed on a new task set.
Here, only the high-level AMDP models need to be recomputed.
This ``PALM-ET given Nav'' agent achieves, as expected, an immediate and statistically significant jumpstart over the algorithm with the same hierarchy, ET, as well as having a shorter time-to-convergence and greater total reward.
Navigation dominates the sample complexity; however, having been learned once, it may be reused.
Thus, transferred knowledge in PALM alleviates the burden of further exploration at that level, allowing the agent to advance more rapidly in learning higher-level AMDPs.
PALM's style of encapsulated, modular transfer is impossible with R-MAXQ and other HRL methods because they learn representations that recurse down to, or ultimately depend on, the transition probabilities, reward function, and ground states specific to the given task MDP. 

We would like to emphasize two major takeaways from these results. First, \textit{subtask independence limits the effects of other models on the one being learned.}
Since each model in PALM is computed independently, the effects of stochasticity are limited.
PALM can, in effect, learn the value functions of these tasks in a more focused manner. In, for example, R-MAXQ, computational efficiency is negatively affected by the need to compute down to the primitive level at every decision point and model update; other HRL algorithms experience similar problems, requiring computation with a direct dependence on lower-level models. 
In a sense, they do not actually abstract away fine-grained temporal details.
The multi-time model of options, for example, requires knowing the joint distribution of all possible time-steps of all options~\cite{abel2019expected}. 

Second, PALM, unlike related approaches, \textit{produces a plan which can be revised}.
Our investigation of R-MAXQ on much larger, more complex MDPs (including the Cleanup tasks) indicates that the intertwined nature of R-MAXQ's computation exacerbates scalability issues.
PALM's independent models inhibit such problems.
Because each AMDP that PALM is learning has its own dedicated model, the model-based exploration that must occur is handled irrespective of child subtasks.
This fact further distinguishes PALM: the L-AMDP subtask models are not a static policy (as with an option) or fragment of a value function (as with a MAXQ subtask), but an encapsulated MDP in its own right. As a result, each PALM subtask can decide independently, within its own confines, what actions to take, abstracting away  details of child subtasks.

%%%%%%%%%%%%%%%%%%%%%%%%%%%%%%%%%%%%%%%%%%%%%%%%%%%%
%% Discussion & Future Work %%
%%%%%%%%%%%%%%%%%%%%%%%%%%%%%%%%%%%%%%%%%%%%%%%%%%%%
\section{Discussion \& Future Work}
% \section{Conclusion}

We introduce Lifted AMDPs as novel, general, and useful representations of behavior. We develop PALM, which performs hierarchical reinforcement learning while eliminating the dependency on human authors: an agent can create a hierarchy where all constituent parts are learned entirely from data. 
PALM has the following novel traits. Deployed to some new task, it learns a transition and reward model for all subtasks in its hierarchy. This hierarchy itself, and any related state abstractions, may be learned from data. Models learned via PALM are transferable to related tasks. We demonstrate the effectiveness of this approach in supporting complex, hierarchical planning without human supervision. 
The properties of independence and modularity make our approach promising for transfer learning, training on a given task and deploying to related ones with reduced retraining and a jumpstart to performance.

PALM's ability to transfer warrants more extensive investigation, and we hope to examine its potential further, especially in cases that require mixed discrete and continuous state space L-AMDPs.
These hierarchies could combine deep and traditional RL, such that subtasks needing perception or control are handled by the former and those requiring more abstract reasoning are addressed by the latter.
Trained wholly in simulation to learn the higher-level symbolic models, models could be transferred over to a continuous, physical domain where only the lowest-level navigational subtasks need to be learned from scratch.

\section*{Acknowledgments}

The material presented here is based in part upon work supported by the National Science Foundation under Grant No. IIS-1813223 and Grant No. IIS-1426452, and by DARPA under grants W911NF-15-1-0503 and D15AP00102.

\bibliographystyle{aaai}
\bibliography{AAAI-WinderJ.5632}

\end{document}